\pdfoutput=1

\documentclass[11pt]{article}

\usepackage[]{EMNLP2022}

\usepackage{times}
\usepackage{latexsym}

\usepackage[T1]{fontenc}

\usepackage[utf8]{inputenc}

\usepackage{microtype}

\usepackage{inconsolata}

\usepackage{amsmath,amsfonts,bm}

\def\eqref#1{equation~\ref{#1}}

\def\1{\bm{1}}

\DeclareMathAlphabet{\mathsfit}{\encodingdefault}{\sfdefault}{m}{sl}
\SetMathAlphabet{\mathsfit}{bold}{\encodingdefault}{\sfdefault}{bx}{n}

\def\gF{{\mathcal{F}}}

\def\gN{{\mathcal{N}}}

\DeclareMathOperator*{\argmax}{arg\,max}

\usepackage{graphicx}
\usepackage{booktabs} %
\usepackage{amsthm,lipsum}

\usepackage{hyperref}
\usepackage{algorithm}
\usepackage[ruled,vlined,algo2e]{algorithm2e}
\usepackage{soul}
\usepackage{caption}
\usepackage{subcaption}

\usepackage[T1]{fontenc}    %
\usepackage{url}            %
\usepackage{amsfonts}       %
\usepackage{nicefrac}       %
\usepackage[export]{adjustbox}
\usepackage{xfrac}
\usepackage{nccmath}
\usepackage{blindtext}
\usepackage{multirow}

\usepackage{float}

\usepackage{makecell}

\newcommand\op[1]{\operatorname{#1}}

\newcommand\lip{CLIP}
\newcommand\ie{\textit{i.e.}, }
\newcommand\ours{\textsc{WALIP}} 
\newcommand\proc{Procrustes}
\newcommand\muse{\textsc{MUSE}}
\newcommand\globe{\textsc{Globetrotter}}
\newcommand\muve{\textsc{MUVE}}
\newcommand\htw{HowToWorld}
\newcommand\update[1]{#1}

\usepackage{etoolbox}
\newcommand{\zerodisplayskips}{%
  \setlength{\abovedisplayskip}{0pt}%
  \setlength{\belowdisplayskip}{0pt}%
  \setlength{\abovedisplayshortskip}{0pt}%
  \setlength{\belowdisplayshortskip}{0pt}}
\appto{\normalsize}{\zerodisplayskips}
\appto{\small}{\zerodisplayskips}
\appto{\footnotesize}{\zerodisplayskips}

\usepackage{siunitx}
\usepackage[textsize=scriptsize]{todonotes}

\usepackage{cleveref}

\let\svthefootnote\thefootnote
\newcommand\freefootnote[1]{%
  \let\thefootnote\relax%
  \footnotetext{#1}%
  \let\thefootnote\svthefootnote%
}

\title{
Utilizing Language-Image Pretraining \\ for Efficient and Robust Bilingual Word Alignment}

\author{Tuan Dinh\footnotemark[1], Jy-yong Sohn, Shashank Rajput, Timothy Ossowski, \\ {\bf Yifei Ming, Junjie Hu, Dimitris Papailiopoulos, Kangwook Lee}  \\
        University of Wisconsin, Madison, WI, USA}

\begin{document}

\maketitle
\freefootnote{$^*$Email: Tuan Dinh (tuan.dinh@wisc.edu)}

\begin{abstract}
Word translation without parallel corpora has become feasible, rivaling the performance of supervised methods. 
Recent findings have shown the improvement in accuracy and robustness of unsupervised word translation (UWT) by utilizing visual observations, which are universal representations across languages.
Our work investigates the potential of using not only visual observations but also pretrained language-image models for enabling a more efficient and robust UWT. 
We develop a novel UWT method dubbed Word Alignment using Language-Image Pretraining (\textit{\ours{}}), leveraging visual observations via the shared image-text embedding space of CLIPs~\citep{radford2021learning}. 
\ours{} has a two-step procedure. 
First, we retrieve word pairs with high confidences of similarity, computed using our proposed \emph{image-based fingerprints}, which define the initial pivot for the alignment.
Second, we apply our \emph{robust Procrustes algorithm} to estimate the linear mapping between two embedding spaces, which iteratively corrects and refines the estimated alignment.
Our extensive experiments show that \ours{} improves upon the state-of-the-art performance of bilingual word alignment for a few language pairs across different word embeddings and displays great robustness to the dissimilarity of language pairs or training corpora for two word embeddings. 
\end{abstract}
\section{Introduction}

Translating words across different languages is one of the long-standing research tasks and a standard building block for general machine translation.
Word translation is helpful for various downstream applications, such as sentence translation~\citep{conneau2017word,hu-etal-2019-domain-adaptation} or cross-lingual transfer learning in language models~\citep{de2020good}.
Unsupervised word translation (UWT) has recently drawn a great deal of attention~\citep{artetxe2017learning,conneau2017word,hartmann2019comparing}, reducing the need for bilingual supervision.

\begin{figure}[t]
\centering
\includegraphics[width=\columnwidth]{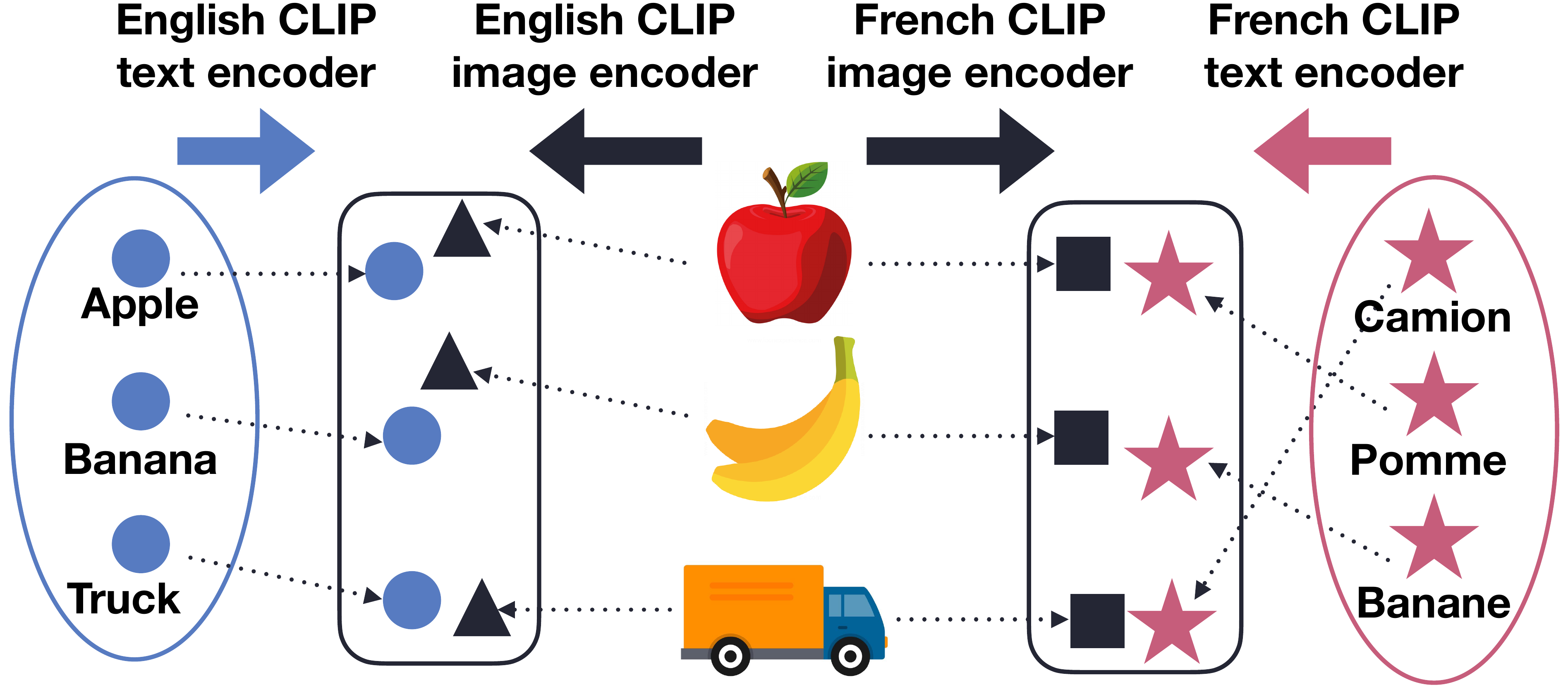}
\vspace{-8mm}
\caption{Conceptual visualization of \ours{} for unsupervised word translation between English and French.
We can connect English and French words in an unsupervised fashion through the shared images.
CLIP models~\citep{radford2021learning} can be used as human simulators to associate words with images.
}\label{fig:concept}
\vspace{-4mm}
\end{figure}

Without any prior knowledge of the languages' connection, aligning their words is non-trivial.
Most works on UWT exploit the structural similarity between continuous word embedding spaces across languages~\citep{mikolov2013exploiting,ormazabal2019analyzing} to learn a linear mapping.
Early works~\citep{smith2017offline,artetxe2017learning,conneau2017word,hoshen2018non,grave2019unsupervised} focus on using only the text data to establish the bilingual alignment and solve the \proc{} problem~\citep{schonemann1966generalized}.
These methods rely on the similarity between pairs of languages and training corpora, thus not working well when the languages or corpora are dissimilar~\citep{sogaard2018limitations,sigurdsson2020visual}. They may also need a large amount of data to achieve good alignments~\citep{sigurdsson2020visual}.

Words can also be connected via the visual world.
Visual similarity provides additional prior knowledge for easing language translation~\citep{mihalcea2008toward}.
Recent works~\citep{sigurdsson2020visual,suris2020globetrotter} demonstrate the promise of using visual information to improve UWT.
However, they mostly require intense joint training for the embedding shared between images or videos with texts of multiple languages.
Moreover, these embeddings are used for translating all words, whereas not every word can be described by images or videos. 
Thus, it is unclear how they are helpful for non-visual words and whether the methods properly utilize topological similarity between word vector spaces~\citep{mikolov2013exploiting}.

\paragraph{Our contributions.} We propose \ours{}
(\textbf{W}ord \textbf{A}lignment with \textbf{L}anguage-\textbf{I}mage \textbf{P}retraining) as a new unsupervised word alignment method that leverages the joint image-text embeddings provided by CLIP~\citep{radford2021learning}.
Fig.~\ref{fig:concept} shows an example inspiring \ours{}.
Consider a conversation between a French and an English speaker. As the English speaker shows an \texttt{apple} image, the French speaker can easily understand and provide its translation as \texttt{pomme}. 
They can similarly pair more words describing simple objects, helping translate more complex words.
This observation inspires us to leverage visual information as the pivot for matching words across languages.
To do so, we use CLIP~\citep{radford2021learning}  to correlate texts and images and construct an image-based word representation, called a \textit{fingerprint}, where each coordinate measures the similarity between the word and an image a diverse image set.
Note that fingerprints share similar merits with the pictorial representation of sentence~\citep{mihalcea2008toward} that represents simple sentences by sequences of pictures.
We use fingerprints to identify initial word pairs.
As not every word can be described by images, we rely on the topological similarity of word vector spaces~\citep{mikolov2013distributed} for the full alignment in the second step, \textit{i.e.},
solving a linear mapping between two spaces using our robust Procrustes algorithm with identified word pairs.

Via extensive experiments, we show that \ours{} is highly effective in bilingual alignment.
We achieve comparable or better performance than the state-of-the-art (SOTA) baselines and close the gap to supervised methods.
For instance, on the Dictionary benchmark~\citep{sigurdsson2020visual} with \htw{}-based word embedding~\citep{miech2019howto100m}, we achieve the SOTA performance on all evaluated pairs (English$\rightarrow$\{French, Korean, Japanese\}), achieving significant accuracy improvements (6.7\%, 2.5\%, and 4.5\%, \emph{cf.} Table~\ref{tab:htw}) over the previous SOTA~\citep{sigurdsson2020visual}.
Our method also displays great robustness to the dissimilarity of language pairs and static word embeddings.
We empirically show the effectiveness of our method through various ablation studies.

\section{Related Works}

\paragraph{Unsupervised word translation (UWT).}
Most UWT methods exploit the structure similarity between word vector spaces across languages~\citep{mikolov2013exploiting} to learn linear mappings.
Early works~\citep{smith2017offline,artetxe2017learning} establish the parallel vocabulary and estimate the mapping by solving the \proc{} problem~\citep{schonemann1966generalized, gower2004procrustes}.
Others study assignment problems and directly solve Wasserstein-Procrustes for the one-to-one word assignment matrix~\citep{zhang2017earth,grave2019unsupervised} or hyper-alignment for multiple languages~\citep{alaux2018unsupervised,taitelbaum-etal-2019-multi}.
Recent works~\citep{zhang2017adversarial,conneau2017word,hoshen2018non} propose to learn the mapping via aligning the embedding distributions with the notable MUSE framework~\citep{conneau2017word} using the adversarial training to achieve high translation performance for multiple pairs.
We use MUSE as our baseline.
While MUSE involves intense training for aligning two embedding spaces,
\ours{} does not require this training by utilizing pretrained CLIP models.

Visual information has been used to improve machine translation~\citep{hewitt2018learning,zhou2018visual,kiros2018illustrative,yang2020using,li2022valhalla}.
Focusing on word translation, MUVE~\citep{sigurdsson2020visual} trains a linear mapping between two embeddings via learning a joint video-text embedding space for pairs with captioned instructional videos.
Globetrotter~\citep{suris2020globetrotter} learns the multilingual text embeddings aligned with image embeddings via contrastive learning.
The learned text embeddings are used for multilingual sentence translation and refined for word translation.
These methods require intense training with a large amount of vision-text data for learning the encoders, while \ours{} only utilizes pretrained embeddings of off-the-shelf CLIP models.
MUVE and Globetrotter are our main baselines.

\paragraph{Language-Vision (LV) models.} 
We  can categorize LV models into two types: single-stream and dual-stream models.
The former feeds the concatenation of text and visual features into a single transformer-based encoder, such as VisualBERT~\cite{li2019visualbert} and ViLT~\cite{kim2021vilt}.
The latter uses separate encoders for text and image and aligns semantically similar features in different modalities with contrastive objectives, such as CLIP~\cite{radford2021learning}, ALIGN~\cite{jia2021scaling}, and FILIP~\cite{yao2021filip}. 
We use CLIP as our language-image pretraining model due to its inference efficiency, high performance, and the availability of pretrained models in multiple languages.
CLIP inspires numerous  works~\cite{zhang2021tip,li2021supervision,zhou2022cocoop} for better data efficiency and task adaptation of LV models.
In this line of work, \citet{zhai2022lit} recently show the feasibility of training multilingual image-text models without parallel corpora by connecting languages via image embeddings.

\section{Problem Setup and Preliminaries}

We formally describe the target problem of unsupervised word alignment and provide two preliminaries to our method: \proc{} and CSLS.

\noindent\textbf{Unsupervised word alignment.~}
We focus on the word alignment (translation) problem: finding the mapping from $A_{\op{dict}}$ to $B_{\op{dict}}$, where  $A_{\op{dict}} = \{a_1, \cdots, a_{n_a} \}$ and $B_{\op{dict}} = \{b_1, \cdots, b_{n_b} \}$ are dictionaries of source language $A$ and target language $B$, with $n_a$ and $n_b$ being the number of words in each dictionary, respectively.
This mapping can be represented by an equivalent index mapping $\pi: [n_a] \rightarrow [n_b]$, \textit{i.e.},
we consider word $a_i$ is mapped (aligned) to word $b_{\pi(i)}$, for $i \in [n_a]$. 
Here, $[n] = \{1, 2, \cdots, n\}$ is defined as the set of positive integers up to a positive number $n$.
Note that we focus on \emph{unsupervised} word alignment in which no ground-truth word pairs $(a_i, b_{\pi(i)})$ are given to the algorithm.
To solve this problem, we assume the access to three ingredients: (1) a large-scale image dataset with $d$ images denoted by $G = \{g_1, \cdots, g_d\}$, (2) a pre-trained monolingual \lip{} model for each language, and (3) static word embeddings~\citep{bojanowski2016enriching,pennington2014glove} for all words in dictionaries.

\label{sec:prelim}
\noindent\textbf{Procrustes problem.~}
Let $X, Y \in \mathbb{R}^{n \times d}$ be matrices of the $d-$dimensional embeddings for $n$ words in the source and target languages.
The Procrustes problem aims to find $W \in \mathbb{R}^{d \times d}$ such that $\lVert XW - Y \rVert_F $ is minimized.
Regularizing $W$ with the orthogonality is found to improve the translation~\citep{xing2015normalized}, where the optimal $W$ is 
\begin{align*}
W^* = \underset{W \in \mathcal{O}_d}{\texttt{argmin}} \|XW - Y\|_F = \texttt{SVD}(Y^TX)
\end{align*}%
where $\mathcal{O}_d$ is the set of $d\times d$ orthogonal matrices and $\texttt{SVD}$ is the singular value decomposition.

\noindent\textbf{CSLS.~} \citet{conneau2017word} proposed Cross-domain Similarity Local Scaling (CSLS) to robustly measure the similarity between words' embeddings.
Given two sets $X = \{x_i\}_{i \in [n_X]}$, $Y = \{y_i\}_{i \in [n_Y]}$ and the number of neighbors $K$, the CSLS of $x_i$ and $y_j$ is defined as 
    $\op{CSLS} (x_i, y_j) = 2 \cos (x_i, y_j) - r_Y (x_i) - r_X (y_j)$
where 
$\cos(\cdot, \cdot)$ is the cosine similarity,
$r_Y(x_i) = \frac{1}{K} \sum_{y_j \in \gN_{Y}(x_i) } \cos( x_i, y_j )$ is the average similarity of $x_i$, and $\gN_Y(x_i)$ is the set of $K$ nearest neighbors of $x_i$ among elements of $Y$.
CSLS performs cross-domain normalization to address the hub phenomenon~\citep{radovanovic2010hubs} of the $K$-nearest-neighbor method in high-dimensional  spaces, which occurs when some vectors are nearest to many vectors while others are isolated.

\section{\ours{}}
\label{sec:walip}
We first provide the high-level idea and then specify each stage of \ours{}.
Algo.~\ref{alg:C_UWT} in Appendix presents the pseudocode for our algorithm. 

\subsection{Method Overview}
\begin{figure}[t]
    \centering
    \includegraphics[width=0.47\textwidth]{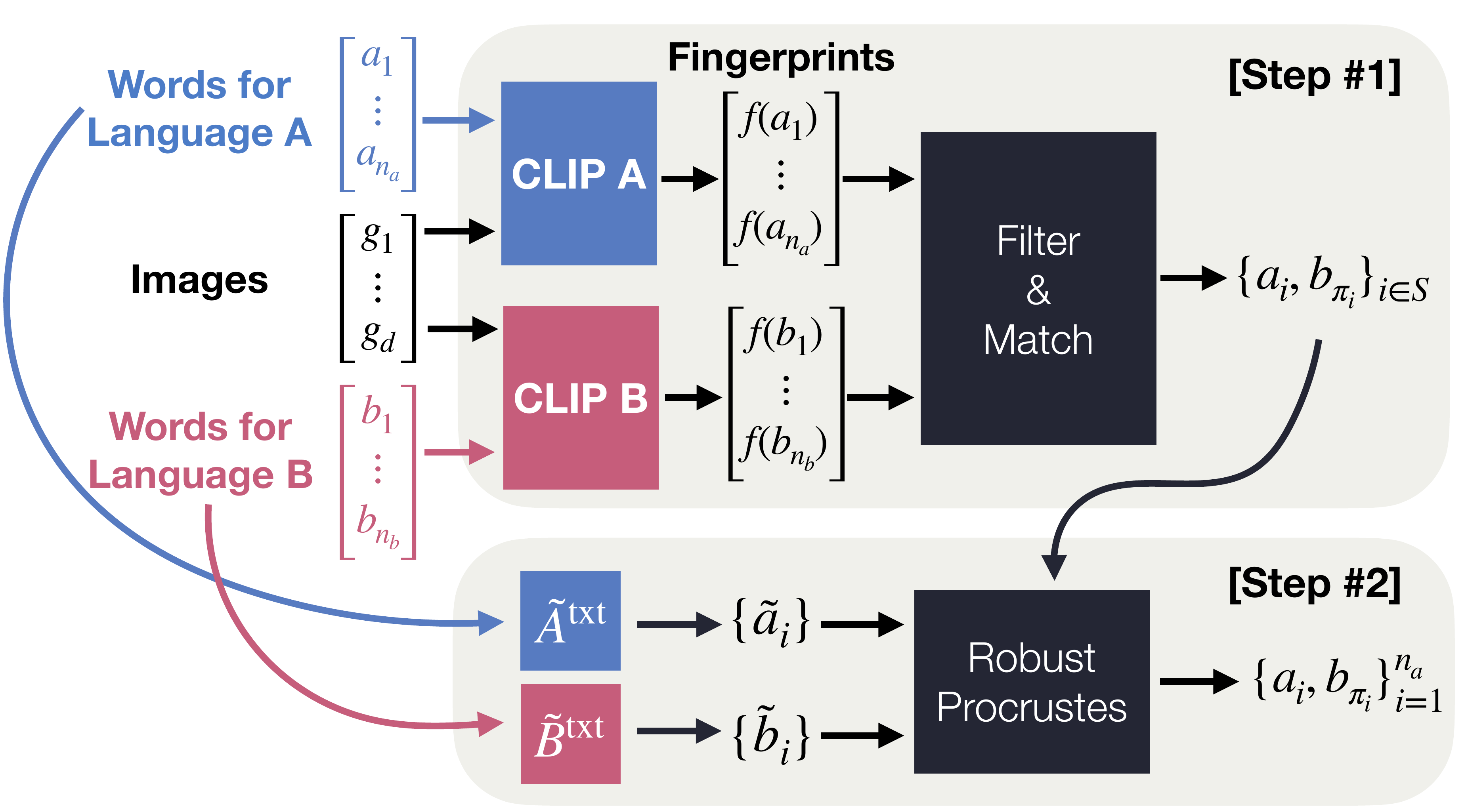}    
    \vspace{-2mm}
    \caption{\ours{} for translating between $n_a$ words $\{a_1, \cdots, a_{n_a}\}$ and $n_b$ words $\{b_1, \cdots, b_{n_b}\}$ in two languages $A$ and $B$. 
    We have access to: (1) a set of $d$ images $\{g_i\}_{i=1}^d$, (2) the CLIP model for each language, 
    and (3) static word embeddings for each language, denoted by $\tilde{A}^{\op{txt}}$ and $\tilde{B}^{\op{txt}}$. 
    In step 1, we build fingerprints $f(a_i)$, $f(b_i)$ defined in~\eqref{eqn:fingerprint} for each word $a_i$, $b_i$.
    We match words whose fingerprints share high similarities, thus having an initial mapping $\pi : [n_a] \rightarrow [n_b]$ pairing $a_i$ and $b_{\pi_i}$ for $i \in S \subseteq [n_a]$.
    In step 2, we use static word vectors and initially matched pairs to solve the linear word mapping with the robust \proc{} algorithm for better alignment.    
    }
    \label{fig:cuwt}
    \vspace{-4mm}
\end{figure}
Our idea is to enable effective and robust word alignment by using (1) the similarity of visual representations of words with similar meanings and (2) the structural similarity of static word embedding spaces across languages. Specifically, we use images to connect similar words in two languages with the aid of CLIP~\citep{radford2021learning}.
However, a na\"ive application of this method only makes sense for visual words such as non-abstract nouns that images can describe.
To map non-visual words, we utilize the topological similarity (\textit{i.e.}, the \textit{degree of isomorphism}) between word vector spaces~\cite{vulic-etal-2020-good}.
Motivated by the existence of a linear association between two static word embeddings of different languages~\citep{ormazabal2019analyzing}, we learn a linear mapping using the robust matching algorithm on identified word pairs.

Fig.~\ref{fig:cuwt} illustrates \ours{} used for aligning  words $\{a_i\}$ and words $\{b_i\}$  in languages $A$ and $B$.
\ours{} has two steps. 
First, it selects pairs $\{a_i, b_{\pi_i}\}$ having similar visual meanings by using each word's fingerprint, defined as the similarity of the word and an image set via \lip{}'s encoders.
Second, it iteratively aligns word embeddings of languages $A$ and $B$, \ie find a linear mapping between two embeddings, using robust \proc{} and the initial pairs identified in the first step.

\subsection{Step 1: Pairing Visually Similar Words using Language-Image Association}\label{sec:step1}

As shown in Algo.~\ref{alg:C_UWT}, our Step 1 pairs words via images. This is available by an image-based fingerprint representation of each word, defined below. 

\subsubsection{Image-based Fingerprints}

We denote the image/text encoder of the \lip{} model for language $A$ as $A^{\op{img}}$ and $A^{\op{txt}}$.  
Similarly, we define $B^{\op{img}}$ and $B^{\op{txt}}$ for language $B$.
The critical advantage of the \lip{} model is the access to the shared embedding space aligning image $g_i$ and its corresponding word ($a_i$ or $b_i$). 
\ours{} utilizes this embedding space of each source/target language to find the bilingual mapping.

\begin{figure}
    \centering
     \begin{subfigure}[h]{0.45\textwidth}
    \includegraphics[width=\textwidth]{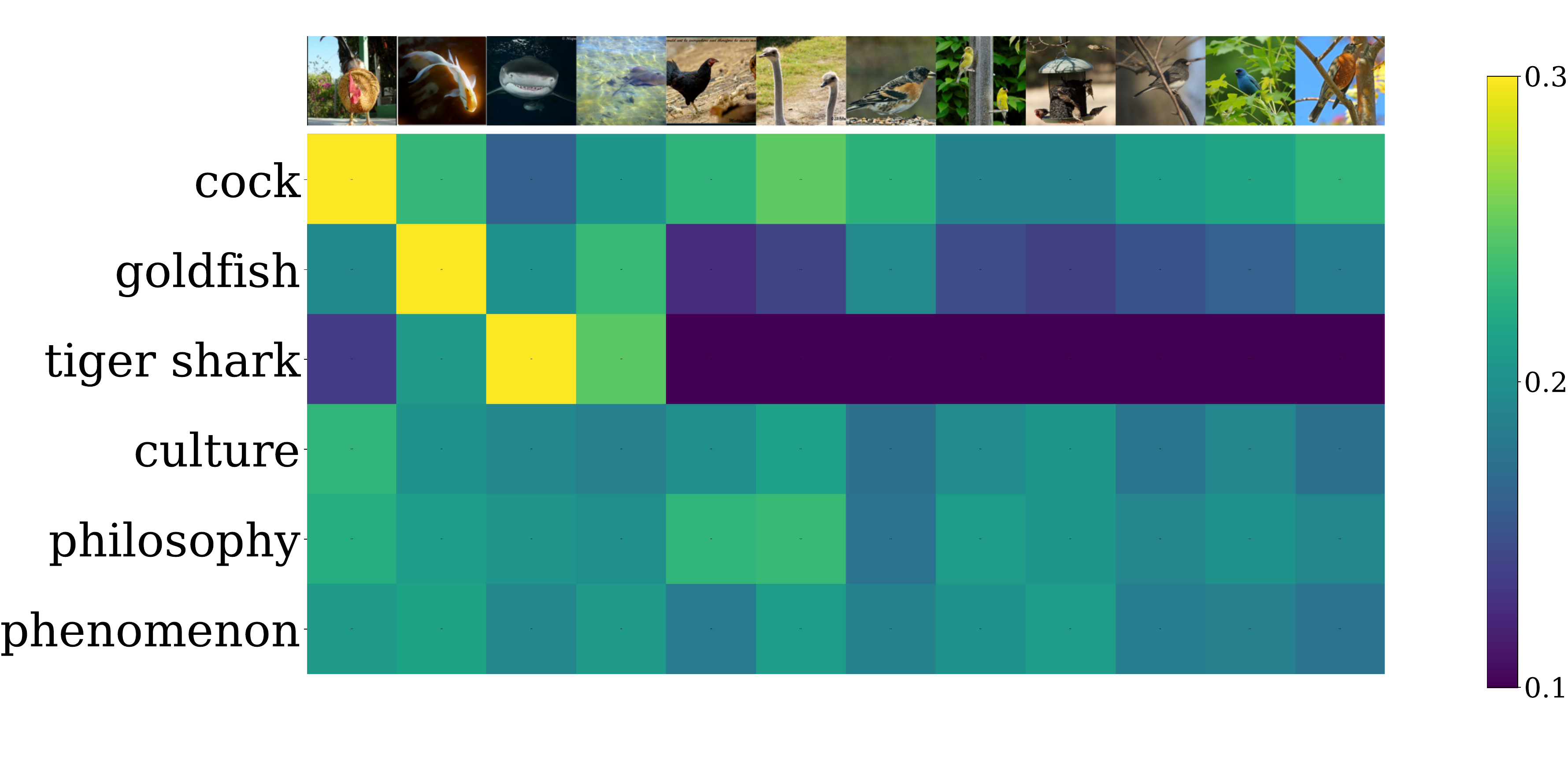}
    \vspace{-9mm}
    \caption{Fingerprints for English words} \label{fig:fingerprint_en}
     \end{subfigure}
     \begin{subfigure}[h]{0.45\textwidth}
    \includegraphics[width=\textwidth]{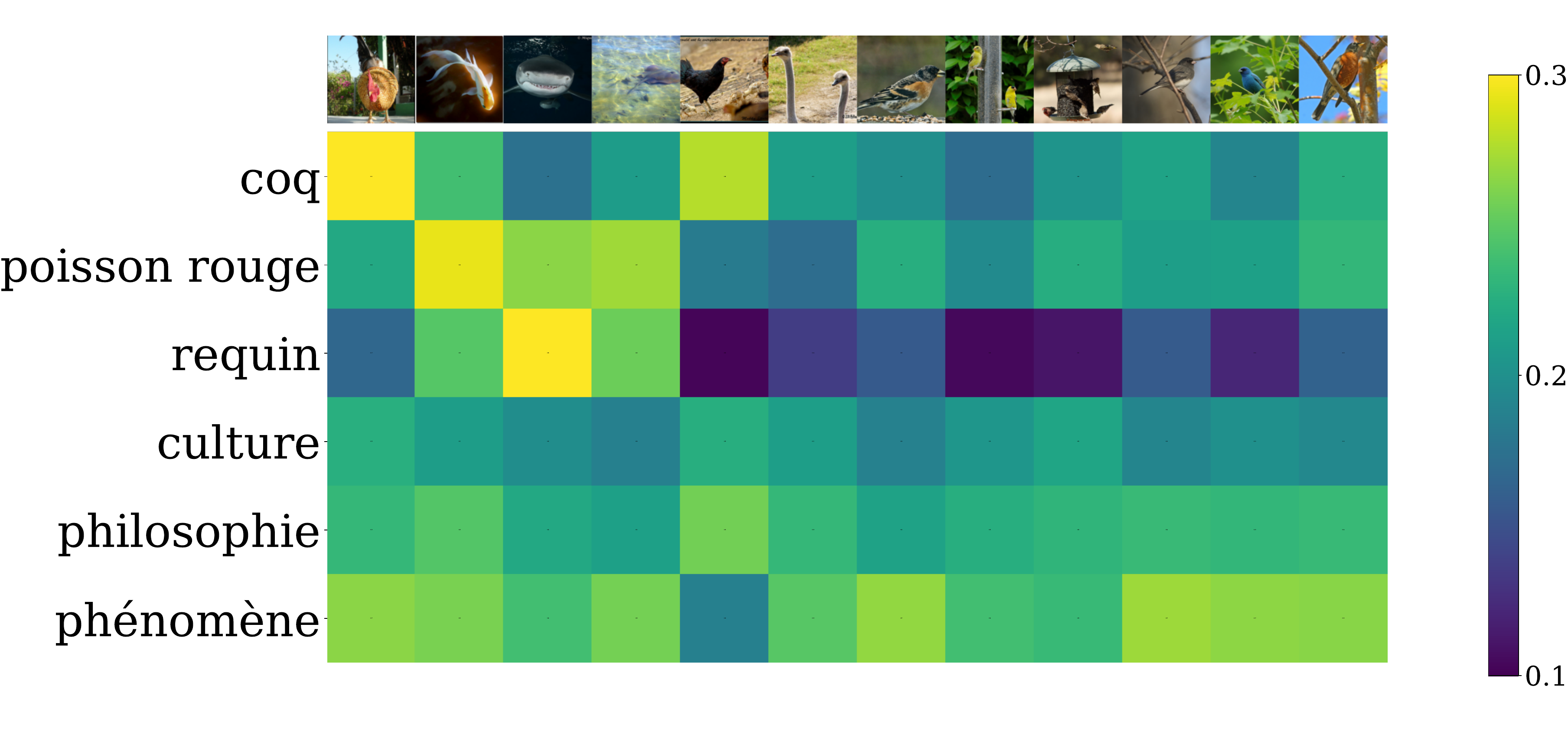}
    \vspace{-9mm}
    \caption{Fingerprints for French words} \label{fig:fingerprint_fr}
      \end{subfigure}
    \vspace{-2mm}
    \caption{Illustration of image-based fingerprints for English words (a) and their translations in French (b). 
    The similarity between each word (inserted in a template such as ``\texttt{A photo of []}'') 
    and all images serves as the fingerprint (each row).       
    Fingerprints of visual words (top three rows) are more distinguishable than abstract words (three bottom rows) and share similar patterns to fingerprints of their French translations.
    }
    \label{fig:fingerprint}
    \vspace{-4mm}
\end{figure}

Given $d$ images $\{g_1, \cdots, g_d\}$, we first define a $d-$dimensional vector (called \textbf{fingerprint}) 
for each word $a_i \in A_{\op{dict}}$ in the source language as 
$f(a_i) = [f_{i,1}^{a}, \cdots, f_{i,d}^{a}]$    
where  $f_{i,j}^{a} = \op{sim}(A^{\op{txt}}(a_i), A^{\op{img}}(g_j))$ is the similarity between the embedding of the $i$-th word and the embedding of the $j$-th image. 
Similarly, we define the fingerprint of each word $b_i \in B_{\op{dict}}$ in the target language as 
$f(b_i) = [f_{i,1}^{b}, \cdots, f_{i,d}^{b}]$    
where $f_{i,j}^{b} = \op{sim}(B^{\op{txt}}(b_i), B^{\op{img}}(g_j))$.
This fingerprint represents a word's similarity to images, according to the embedding space of pretrained \lip{} models. 
We denote the fingerprint of the $i$-th word in the dictionary of a language $l \in \{a, b\}$ as 
\begin{align}\label{eqn:fingerprint}
f(l_i) = [f_{i,1}^{l}, \cdots, f_{i,d}^{l}].    
\end{align}%
Figs.~\ref{fig:fingerprint_en},~\ref{fig:fingerprint_fr} show examples of English and French fingerprints.
Here, we measure the similarity of each word with 12 images from ImageNet~\citep{deng2009imagenet}, obtaining a 12-dim vector.
The top three rows of each figure are fingerprints for visual words (\texttt{cock, goldfish, tiger shark}), and the bottom rows are of abstract words (\texttt{culture, philosophy, phenomenon}).
Unlike visual words, fingerprints of abstract words are more uniform  (similar values for most coordinates), \textit{i.e.,} they are \emph{not} distinguishable.
Note that fingerprints of each English-French pair of visual words \{(\texttt{cock, coq}), (\texttt{goldfish, poisson rouge}), (\texttt{tiger shark, requin})\} share highly similar patterns.

\subsubsection{Identifying Pivot Pairs}
Consider two visual words $a_i, b_j$ in two languages with similar meanings (\textit{e.g.}, $a_i = \texttt{``tiger shark''}$ and $b_j = \texttt{``requin''}$ in Fig.~\ref{fig:fingerprint}). 
For a given image set, fingerprints of the two words would be similar, \textit{i.e.}, $f(a_i) \approx f(b_j)$ as shown in Fig.~\ref{fig:fingerprint}, allowing the use of fingerprint similarity for word translation.

\paragraph{Keeping only visually aligned words.}
Recall that fingerprints are meaningful for visual words only, as observed in Fig.~\ref{fig:fingerprint}.
Motivated by this observation, we focus on words well represented by a set of images. 
Specifically, for the $i$-th word $l_i$ in language $l \in \{a, b\}$, we compute the maximum similarity value $f_{i,\op{max}}^{(l)} = \max_j f_{i,j}^{(l)}$ within the corresponding fingerprint $f(l_i) = [f_{i,1}^{(l)}, \cdots, f_{i,d}^{(l)}]$.
Then, for each language $l \in \{a, b\}$, we keep the set of words $S_l$ having the maximum similarity beyond the median. 
To focus on components with high similarity, we sparsify fingerprints by eliminating values below the $0.9^{th}$-quantile and normalize the vectors.
This revised fingerprint allows us to focus on images highly similar to the given word. 

\paragraph{Selecting pairs with high similarity.} %
For source words $\{a_i\}_{i \in S_a}$ and target words $\{b_j\}_{j \in S_b}$, we measure the similarity of fingerprints $f(a_i)$ and $f(b_j)$ using CSLS (Sec.~\ref{sec:prelim}). Recall that our goal is to find a mapping $\pi: [n_a] \rightarrow [n_b]$ indicating that the word $a_i$ is translated to $b_{\pi(i)}$, and we want to map $a_i$ to $b_j$ having similar fingerprints. 
Based on the similarity score $c_{i,j} = \op{CSLS}(f(a_i), f(b_j))$ for $i \in S_a$ and $j \in S_b$, we set $\pi(i) = \argmax_{j} c_{i,j}$, giving us an initial set of word pairs, where two words in each pair are visual words and share highly similar fingerprint patterns. 
See algorithms~\ref{alg:filter-words} and \ref{alg:csls_nn} for pseudocodes of word filtering and pair selection.

\subsection{Step 2: Iteratively Learning the Mapping with Robust \proc{}}\label{sec:step2}
\label{sec:method_step2}
In Step 1 of \ours{} in Algo.~\ref{alg:C_UWT},
we have identified the initial word mapping $\pi$ on visual words.
In Step 2, we learn and fine-tune $\pi$ on the whole dictionaries using linear mapping $W^{\star}$ between \textit{static} word embeddings
of two languages, learned by iteratively applying our robust Procrustes algorithm (Algo.~\ref{alg:robust_procrustes}).
We first explain Algo.~\ref{alg:robust_procrustes} -- the building block of Step 2 in Sec.~\ref{sec:robust_proc}, and then explain how this algorithm allows us to learn $\pi$ in Sec.~\ref{sec:iterative}.

\begin{algorithm}[!htbp]
\footnotesize
\caption{\texttt{Robust-Procrustes}}
\label{alg:robust_procrustes}
\SetKwInput{Input}{Input}
\SetKwInput{Output}{Output}
\Input{
    Vectors $X, Y \in \mathbb{R}^{n\times d}$ \\
}
\Output{
    Linear mapping $W^*\in \mathbb{R}^{d\times d}$
}
{
Set $\epsilon=0.001$, $M = 5$ \\
Initial mapping $W_0 = \texttt{Procrustes}(X, Y)$ \\
}
\For{$m \in \{1, \cdots, M\}$}{
{
$\alpha_i \leftarrow \frac{1}{\|y_i - W_{k-1}x_i\|^2 + \epsilon} \quad \text{for}~~ i\in[n]$\\
$\alpha_i \leftarrow \alpha_i/\max_{j\in[n]}{\alpha_j}$ \\
$D \leftarrow \texttt{Diag}(\alpha_1^{1/2}, \dots, \alpha_n^{1/2})$ \\
$W_m \leftarrow \texttt{Procrustes}(DX, DY)$
}
}%
$W^{\star} \leftarrow W_M$
\end{algorithm}

\subsubsection{Error-Weighting Robust Procrustes}
\label{sec:robust_proc}
The initial word pairs identified in Step 1 are obtained in an unsupervised manner with potentially many mismatched pairs. Thus directly applying the existing Procrustes algorithm (Sec.~\ref{sec:prelim}) to these pairs may lead to an incorrect linear mapping $W$.

We introduce a robust matching algorithm (Algo.~\ref{alg:robust_procrustes}) to eliminate the mismatched pairs and learn the mapping from the correct ones.
Inspired by the existing robust \proc{} algorithms~\citep{groenen2005improved}, we assign small weights to incorrect pairs and large weights to correct pairs.
Given a word embedding matrix $X$ and its aligned counterpart $Y$,
we first apply the Procrustes to learn the initial $W_0$.
We then measure the error of $W_0$ on each word pair $(x,y)$ by the residual $r(x, y) = \|y-W_0x\|_2$.
Since the pair is likely to be correct when the residual is small, 
we use $\alpha(x, y) = \sfrac{1}{r(x,y)}$ as the weight of the pair. 
Then, we apply \proc{} on these weighted pairs to obtain a new mapping $W_1$.
We repeat this process a few times
to achieve a stable linear mapping $W^{\star}$.

\subsubsection{Iteratively Updating the Word Alignment $\pi$ and Linear Mapping $W^{\star}$}
\label{sec:iterative}
In Step 2 of \ours{}, we iteratively apply two procedures: first, we update linear mapping $W^{\star}$ by applying the robust \proc{} on identified pairs, and second, we update the word mapping $\pi$ using $W^{\star}$ and the pair selection algorithm (Algo.~\ref{alg:csls_nn}).

The first phase is described in Sec.~\ref{sec:robust_proc}.
In the second phase, we transform each source vector $x_i$ into $W^{\star}x_i$ in the target embedding space and apply the $k$-nearest-neighbor (NN) on this space.
We update $\pi$ using Algo.~\ref{alg:csls_nn} in the following manner: retrieving $k>1$ candidate target words for each source word and choosing candidates having the similarity (with source word) higher than a threshold $q$.
For the updated $\pi$, we measure the Euclidean distance between paired vectors as the validation loss and repeat the two procedures (update $W^{\star}$ and $\pi$) until the validation loss is convergent.
In this process, two hyperparameters $q$ and $k$ are initialized with high values and gradually decayed at each update step of $\pi$.
Once the validation loss converged, we obtain the final mapping $\pi$ by applying Algo.~\ref{alg:csls_nn} with $k=1$ and $q=0$.

Step 2 is crucial to achieving high translation performance from initial mapping.
While sharing similar merits to ours, the refinement procedure~\citep{conneau2017word} is only used for marginally improving upon a high-accuracy linear mapping $W$.

\subsection{Advantages of \ours{}}
First, \ours{} is \textit{computationally efficient}, especially compared to MUSE, MUVE, and \globe{}.
With pretrained CLIPs, our first step (Sec.~\ref{sec:step1}) requires no extra training for pivot pair matching, while Step 2 (Sec.~\ref{sec:step2}) involves a few matrix computations.
Second, \ours{} is more \textit{robust to language dissimilarity}. Assuming well-trained CLIPs, fingerprints of words having similar meanings are intuitively similar across languages as they all represent the same visual correlation to the same image set.
Thus, fingerprints improve the robustness of pivot matching, especially for dissimilar languages.
This may not be the case for methods only using static word embeddings~\citep{sogaard2018limitations}.
Finally, our image-based fingerprint provides an \textit{interpretable  representation} of words.
\section{Experiments}
\label{sec:exp}
\update{
We evaluate \ours{} on bilingual alignment tasks. %
Sec.~\ref{sec:exp_performance} compares \ours{} and baselines in multiple language pairs. The following sections provide additional experimental results that either highlight the benefits of \ours{} or help understand the component that enables the high performance of \ours{}. 
Our code is available at \url{https://github.com/UW-Madison-Lee-Lab/walip}.
}

\subsection{Settings} 
\label{sec:exp_setting}

\paragraph{\ours{} setting.}
We use publicly available pretrained CLIPs for English, Russian, Korean, and Japanese.
For other languages, we fine-tune English CLIP models on Multi30K~\citep{W16-3210,elliott-EtAl:2017:WMT} and MS-COCO variants~\citep{lin2014microsoft,IJCOL:scaiella_et_al:2019,ms_coco_es}. 
For making CLIP prompts, we convert single words to sentences using prompt templates suggested in~\citep{radford2021learning}.
We apply the prompt-ensemble technique with 2--7 prompts for each word and use their average as word embeddings.
\update{
To make the fingerprints, we use a set of 3000 images from  ImageNet~\citep{deng2009imagenet} by default.
See Sec.~\ref{sec:ab_imageset} for our detailed evaluation.
For the static word embedding, we use \htw{} (HTW)-based Word2Vec~\citep{sigurdsson2020visual} and Wiki-based Fasttext embeddings~\citep{bojanowski2016enriching}.
}

\paragraph{Evaluation.} We evaluate methods on the \textit{Dictionary} datasets~\citep{sigurdsson2020visual}, which are test sets used in the MUSE benchmark~\citep{conneau2017word}.
Each dictionary is a set of translation pairs where each word in the source language may have multiple translations in the target language.
We report {recall@n} used in~\cite{sigurdsson2020visual}, which presents the fraction of source words correctly translated.
A retrieval is correct for a given query if at least one of $n$ retrieved words is the correct translation.
By default, we report {recall@1}, which is equivalent to {precision@1}, and the accuracy used in~\citep{conneau2017word}.

\paragraph{Baselines.} 
Our baselines include the video-grounding method \textit{MUVE}~\citep{sigurdsson2020visual} and the image-grounding method \textit{Globetrotter}~\citep{suris2020globetrotter}.
We also compare our method with two versions of the text-only method \textit{MUSE}~\citep{conneau2017word}: the default one trained on the Dictionary dataset (with 1.5K--3K words per dictionary), and the other one trained on the MUSE training data (with 200K words per dictionary); we call the latter one as \textit{MUSE (extra-vocabulary)}.
We also consider a simple baseline using CLIP, denoted by \textit{CLIP-NN}, which performs $1$-nearest neighbor ($1$-NN) based estimation on the embedding spaces of two \lip{} models: we first find the image nearest to the source word, and then find the target word nearest to the image found in the first step. 
For measuring {recall@n} of this baseline, we replace $1$-NN with $\lceil\sqrt{n}\rceil$-NN.

We also test three variants of our method by making changes in Step 1:
\textit{\ours{} (clip-text in Step 1)} which replaces fingerprints with CLIP-based text embeddings, \textit{\ours{} (substring matching)} which replaces the initial matching by selecting pairs sharing the longest common substrings, and \textit{\ours{} (character mapping)} which improves \textit{substring matching} by first applying letter counting~\citep{ycart2012letter} to map two languages' character sets.
We also test two variants that replace the static word embeddings used in Step 2 with CLIP-based text embeddings (denoted by \textit{\ours{} (clip-text in Step 2)}) or fingerprints (denoted by \textit{\ours{} (fingerprint in Step 2)}).
Further details are in Appendix~\ref{sec:app_exp_setting}.

\subsection{How Well Does \ours{} Perform Bilingual Word Alignment?}
\label{sec:exp_performance}

Tables~\ref{tab:wiki@1},~\ref{tab:htw} show our evaluation of bilingual alignment using Wiki-based and HTW-based embeddings on the Dictionary datasets.

\begin{table*}[!htbp]
    \footnotesize
    \renewrobustcmd{\bfseries}{\fontseries{b}\selectfont}
    \sisetup{detect-weight,mode=text,group-minimum-digits =4}
    \caption{
    Comparing bilingual alignment methods on Wiki-based word embedding. We report recall@1 on the Dictionary dataset. 
    \ours{} achieves SOTA performance in many pairs, close to the supervision.
    \textit{\citep{sigurdsson2020visual} do not report results of MUVE in this setting and \globe{} uses its learned word embeddings.}
    }
    \vspace{-2mm}
    \resizebox{\textwidth}{!}
    {
    \centering
        \begin{tabular}{ccSSSSSSSS}
        \hline
        \toprule[1pt]
        & Method & {En$\rightarrow$Ko}& {En$\rightarrow$Ru} & {En$\rightarrow$Fr} & {En$\rightarrow$It} & {En$\rightarrow$Es} & {En$\rightarrow$De} & {Es$\rightarrow$De} &{It$\rightarrow$Fr} \\
        \hline
        \multirow{5}{*}{Text-only} & (Upper bound) Supervision   &69.1  &85.5& 93.5& 92.1& 93.3& 92.5& 91.5&95.1 \\
        \cline{2-10}
        &\muse{} (extra-vocabulary)       & 59.3 &  \bfseries 83.0& \bfseries 92.5 & \bfseries 91.6& \bfseries 93.0& \bfseries 92.5& \bfseries 89.1 &\bfseries 94.5\\
        &\muse{} & 2.8&  65.9 & 84.5 & 84.9 & 85.1& 73.6& 83.0&92.3\\
        & \ours{} (substring matching) & 0.2&  0.0& 92.0& 90.3& 92.0& 92.1 & 88.7 & 94.3\\
        &\ours{} (character mapping) & 0.2 & 5.0& 90.9& 0.1 & 0.1 & 0.3 & 0.5 & 0.5\\
        \hline
        \multirow{6}{*}{Text-Image} & \textsc{CLIP-NN}& 2.5&  9.4& 1.3 & 10.5& 8.2& 7.1& 7.3&6.5\\ %
        & \globe{}    & 0.1& 4.0 & 52.3 & 50.1& 46.4& 46.8& 38.3&49.3 \\
        \cline{2-10}
        &\ours{} (clip-text in Step 1) & 0.3&  0.0& 58.9& 79.4& 56.2 & 50.8 & 46.5 & 52.5  \\
        &\ours{} (clip-text in Step 2) & 0.2 &  15.7& 59.3& 59.1& 59.1& 52.3& 46.8& 52.1\\
        &\ours{} (fingerprint in Step 2) & 0.2& 0.5& 31.3& 39.0& 32.6& 31.3& 34.7&43.3\\
        &\ours{}    &\bfseries 62.3 &  82.7& \bfseries 92.6  & 90.7 & 92.2& \bfseries 92.6& \bfseries 89.2& \bfseries 94.5 \\
        \bottomrule[1pt]
        \end{tabular}
        }
    \label{tab:wiki@1}
    \vspace{-3mm}
\end{table*}

\paragraph{Wiki-based embeddings.} In Table~\ref{tab:wiki@1}, \ours{} achieves comparable or the best performances in most cases among unsupervised methods, attaining relatively small gaps to the full supervision.
Specifically, \ours{} achieves SOTA on five pairs, especially for En$\rightarrow$Ko,  where \ours{} outperforms others with large margins. 
For the baselines using visual information, we outperform \globe{} and all variants of \ours{} across all pairs.
Note that \muve{} only reports recall@10 for En$\rightarrow$Fr as $82.4$, far below ours ($97.5$).
Compared to the version of \muse{} with extra vocabularies, \ours{} achieves comparable scores in most cases and outperforms in En$\rightarrow$Ko. 
The score gaps between the two methods are larger in Table~\ref{tab:htw}, as described in the next paragraph. 
It is worth mentioning that this version of \muse{} needs a large number of extra vocabularies for training while \ours{} directly performs on the test dictionaries. 
Moreover, most baselines (except CLIP-NN) require intense training for aligning embedding spaces, while \ours{} needs a few matrix computations.

\begin{table}[t]    
    \footnotesize
    \renewrobustcmd{\bfseries}{\fontseries{b}\selectfont}
    \sisetup{detect-weight,mode=text,group-minimum-digits =4}
    \caption{Comparing bilingual alignment methods on HTW-based embedding. 
    \ours{} achieves highest {recall@n} scores on Dictionary dataset across all pairs. }
    \vspace{-2mm}
    \centering
    \resizebox{\linewidth}{!}
    {
        \begin{tabular}{cSSSSSS}
        \hline
        \toprule[1pt]
        \multirow{2}{*}{Method} & \multicolumn{2}{c}{En$\rightarrow$Fr} & \multicolumn{2}{c}{En$\rightarrow$Ko} & \multicolumn{2}{c}{En$\rightarrow$Ja}\\
        \cline{2-7}
        {}& {R@1}  & {R@10} & {R@1} & {R@10} & {R@1} & {R@10}\\
        \hline
        (Up.) Sup.  & 57.9 & 80.1 & 41.8 & 72.1  & 41.1&68.3\\
        \hline
        \muse{} (extra.)  & 26.3& 42.3 & 11.8  & 23.9 & 11.6& 23.5\\
        \muse{} & 0.8 & 6.6 & 0.3 & 3.1 & 0.3 & 2.5 \\
        \muve{}     & 28.9& 45.7 & 17.7 & 33.4 & 15.1& 31.2\\
        \hline
        \ours{} (substr.)    & 35.5& 56.0 &  0.0 & 0.2 & 0.3 & 2.1\\
        \ours{}    & \bfseries 35.6& \bfseries 56.2 & \bfseries 20.2 & \bfseries 42.4&\bfseries 19.6 & \bfseries 41.0\\
        \bottomrule[1pt]
        \end{tabular}
    }
    \label{tab:htw}
    \vspace{-4mm}
\end{table}

\paragraph{HTW-based embeddings.}
Following~\cite{sigurdsson2020visual}, we test for three language pairs (En$\rightarrow$\{Fr, Ko, Ja\}).
Table~\ref{tab:htw} compares \ours{} with MUVE and the baselines that perform well in Table~\ref{tab:wiki@1}. 
Results of \muse{} (extra.) and MUVE are from~\citep{sigurdsson2020visual}.
\ours{} outperforms other unsupervised baselines with large margins, achieving the SOTA for all pairs, with the {recall@1} gaps to the second-best method (MUVE) being $6.7, 2.8$, and $4.5$ for En$\rightarrow$\{Fr, Ko, Ja\}, respectively.
\ours{} also outperforms the substring matching variant on En$\rightarrow$\{Ko, Ja\}.

\paragraph{The performance on dissimilar language pairs.} 
For both embedding types, \ours{} works relatively well regardless of the similarity of language pairs. In contrast, most baselines do not perform well on a few or all dissimilar pairs (En$\rightarrow$\{Ko, Ja, Ru\}). We expect that the low performance of the \emph{substring matching} method partly comes from the dissimilarity of alphabets in such pairs.

\subsection{Robustness against the Dissimilarity of Static Word Embeddings}
\label{sec:exp_robust_embeddings}

\begin{table}[t]
    \footnotesize
    \renewrobustcmd{\bfseries}{\fontseries{b}\selectfont}
    \sisetup{detect-weight,mode=text,group-minimum-digits =4}
    \caption{Comparing methods when static word embeddings of source and target languages are trained on different corpora. We report {recall@1} on En$\rightarrow$Fr translation evaluated on Dictionary dataset. 
    \ours{} outperforms other baselines across two settings. 
    }
    \vspace{-2mm}
    \centering
        \begin{tabular}{cSS}
        \hline
        \toprule[1pt]
        Method & {Wiki-HTW} & {HTW-Wiki}\\
        \hline
        \muse{} (extra.)       & 0.3 & 0.3\\
        \muse{} & 0.3 & 0.2 \\
        VecMap    & 0.1 & 0.1\\
        \muve{}     & 32.6 & 41.2\\
        \hline
        \ours{}    & \bfseries 34.3& \bfseries 60.0\\
        \bottomrule[1pt]
        \end{tabular}
    \label{tab:robustness@10}
    \vspace{-2mm}
\end{table}

Following~\citep{sigurdsson2020visual}, we evaluate \ours{} when the static word embeddings of source and target languages come from different training corpora: Wiki and HTW corpora.
We also compare with \textit{VecMap}~\citep{artetxe2017learning}, the baseline used in the MUVE paper. 
Table~\ref{tab:robustness@10} compares \ours{} with MUSE variants, VecMap, and MUVE on En$\rightarrow$Fr.\footnote{MUVE only provides results for the English-French pair.}
\ours{} and MUVE are more robust to the dissimilarity of word embeddings than MUSE variants and VecMap. 
In addition, \ours{} outperforms MUVE on both settings.
For instance, on the Wiki-HTW setting, recall@1 of \ours{} is $60\%$ while that of MUVE is $41.2\%$.

\subsection{Can We Reuse CLIP Models Trained on English Texts for Other Languages?}
\label{exp:zeroshot}
\begin{figure}
    \centering
    \includegraphics[width=0.45\textwidth]{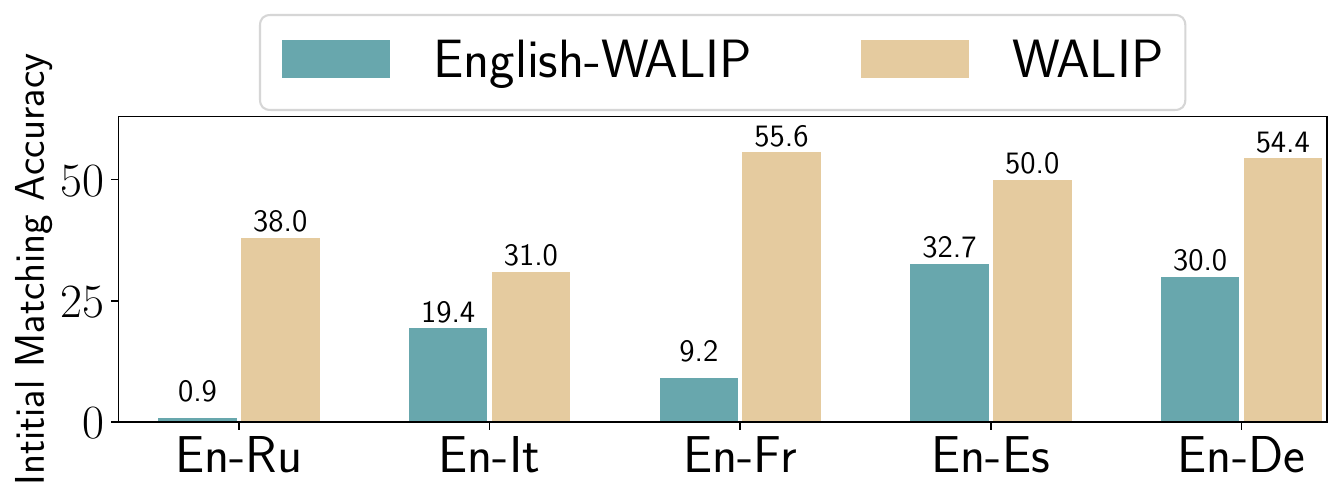}
    \includegraphics[width=0.46\textwidth]{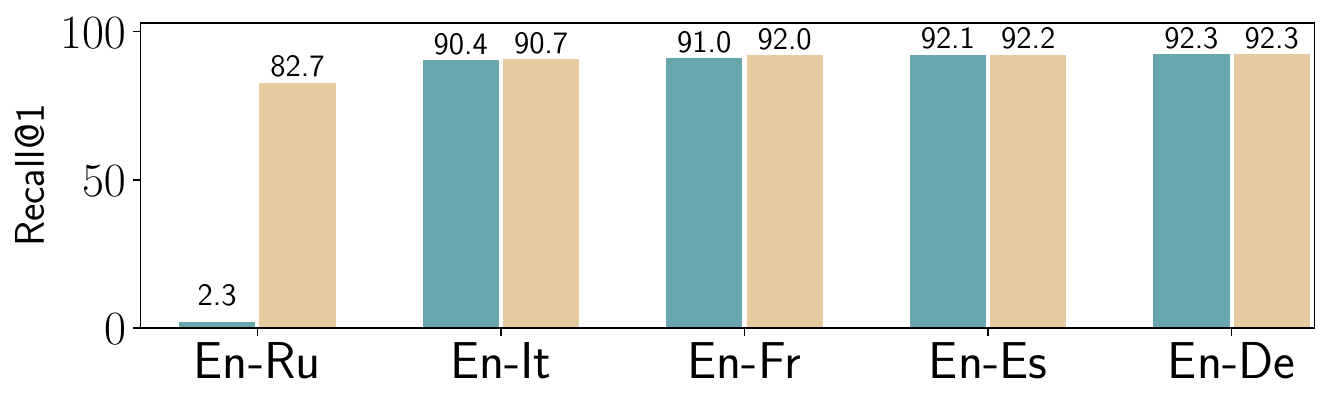}
    \vspace{-3mm}
    \caption{Zero-shot cross-lingual transfer. 
    We observe the following when we replace the original CLIPs (yellow) with English-CLIP (cyan).
    Top: The initial matching accuracy drops for all pairs. Bottom: The final recall score becomes nearly $0$ for the dissimilar pair (En$\rightarrow$Ru) but remains mostly the same for other pairs.}
    \label{fig:ablation_clip_transfer}
    \vspace{-2mm}
\end{figure}

Large-scale language models exhibit the strong ability of cross-lingual zero-shot transfer~\citep{hu2020xtreme}.
We investigate whether \ours{} can utilize a CLIP model trained on English texts (English-CLIP) for other languages.
Intuitively, this is probably doable when the other language uses the same alphabet (and the same tokenizer).
Here, we use the English-CLIP model to obtain fingerprints for all languages, resulting in a new version of \ours{}, denoted \textit{English-\ours{}}.
Here, our static word embeddings are Wiki-based Fasttext embeddings.
As shown in Fig.~\ref{fig:ablation_clip_transfer}, using English-\ours{} causes drops in initial matching accuracies, which measure the precision of mapping on selected pairs.
However, these drops only affect the translation performance of languages dissimilar to English (\textit{e.g.}, Russian)  and do not significantly affect the ones similar to English, \textit{i.e.}, the {recall@1} remains mostly the same for En$\rightarrow$\{It, Fr, Es, De\}.
Thus, English-CLIP can be used in \ours{} framework for languages similar to English, reducing the need for training their new CLIP models.

\subsection{\ours{} on Different Word Types}
\label{exp:word_type}
\begin{table}[t]    
    \footnotesize
    \sisetup{detect-weight,mode=text,group-minimum-digits=4}
    \caption{The percentage (\%) of each word class in the Dictionary dictionaries. Each class of abstract and concrete nouns accounts for approximately $4\%$ of words, with the total nouns being nearly $50\%$ of words.}
    \vspace{-2mm}
    \centering
    \resizebox{\linewidth}{!}
    {
        \begin{tabular}{cSSSS}
        \hline
        \toprule[1pt]
        \multirow{2}{*}{Dict.} & \multicolumn{3}{c}{Noun} & {\multirow{2}{*}{Others}}\\
        \cline{2-4}
        {} & {Abstract}  & {Concrete} & {Non-ID} & {}\\
        \hline
        {En$\rightarrow$Ru}  & 3.8 & 4.3 & 39.5 & 52.4\\
        \hline
        {En$\rightarrow$It} & 3.9 & 3.8 & 38.9 & 53.4\\
        \bottomrule[1pt]
        \end{tabular}
    }
    \label{tab:ab_word_percent}
    \vspace{-3mm}
\end{table}

\begin{table}[t]    
    \sisetup{detect-weight,mode=text,group-minimum-digits=4}
    \caption{{Recall@1} ($\uparrow$) of each word type reported on each step of \ours{}. In the early stages, concrete nouns obtain the highest scores in both  dictionaries. After step 2, abstract and concrete nouns share more comparable scores, higher than scores of non-noun words. }
    \centering
    \vspace{-2mm}
    \resizebox{\linewidth}{!}
    {
        \begin{tabular}{cSSSSS}
        \hline
        \toprule[1pt]
        \multirow{2}{*}{Dict.} &  {\multirow{2}{*}{Step (Iter.)}}  & \multicolumn{3}{c}{Noun} & {\multirow{2}{*}{Others}}\\
        \cline{3-5}
        & & {Abstract}  & {Concrete} & {Non-ID} & \\
        \hline
        \multirow{3}{*}{En$\rightarrow$Ru}  & {\#1} & 7.0 & 47.6& 9.8&5.8\\
        \cline{2-6}
         &  {\#2 (first)} & 40.4&66.2 &42.2 &23.7\\
          & {\#2 (last)} & 86.0& 86.2& 86.0& 78.8\\
        \hline
        \multirow{3}{*}{En$\rightarrow$It}  &{\#1}  & 3.3& 35.1& 13.2& 12.9\\
        \cline{2-6}
            & {\#2 (first)}  & 72.9& 77.2& 68.0& 55.1\\
          & {\#2 (last)} & 96.6& 94.8& 92.0& 89.3\\
        \bottomrule[1pt]
        \end{tabular}
    }
    \label{tab:ab_word_acc}
    \vspace{-4mm}
\end{table}
\update{
In this section, we check how the performance of \ours{} changes for different types of words. 
We categorize words into 4 classes: abstract nouns (\textit{e.g.}, beauty), concrete nouns (\textit{e.g.}, computer), non-identified nouns (\textit{e.g.}, Copenhagen), and non-noun (\textit{e.g.}, pretty).
We use spaCy noun parser\footnote{https://spacy.io} to detect nouns and then use lists of popular English abstract and concrete nouns\footnote{englishvocabs.com/nouns/1000-concrete-and-abstract-nouns-examples, onlymyenglish.com/list-of-abstract-nouns} to match their classes.
We denote the unmatched nouns as non-identified (non-id).
Table~\ref{tab:ab_word_percent} reports the percentage of each class in the En$\rightarrow$\{Ru, It\} dictionaries.
Nearly $47\%$ of words are nouns, with approximately $8\%$ of words being abstract or concrete nouns.}

\update{
Table~\ref{tab:ab_word_acc} reports {recall@1} scores for all word classes after the initial matching (Step 1 in Sec.~\ref{sec:step1}) and after the first and the last iterations of linear mapping (Step 2 in Sec.~\ref{sec:method_step2}).  
We use the Wiki-based Fasttext embeddings for static word embeddings.
After completing step 1 and the first iteration of step 2, concrete nouns have the highest scores.
Note that the score gap between concrete and abstract nouns on En$\rightarrow$Ru is more than $40\%$ after step 1.
This indicates that the initial matching using fingerprints works better with concrete nouns.
After completing step 2, the scores are improved for all classes, where scores of abstract and concrete nouns become more comparable, \textit{e.g.}, $86.0, 86.2$ on En$\rightarrow$Ru.
Note that nouns have much higher {recall@1} than non-noun words.
These results show that step 2 improves the matching for all word types, especially for nouns.
}

\subsection{Ablation Study}\label{sec:ablation}
\label{sec:exp_ablation}
We perform ablation studies using the Wiki-based Fasttext embedding and the Dictionary dataset.

\subsubsection{Effect of Fingerprints}
\begin{figure}[t]
    \centering
    \includegraphics[width=0.48\textwidth]{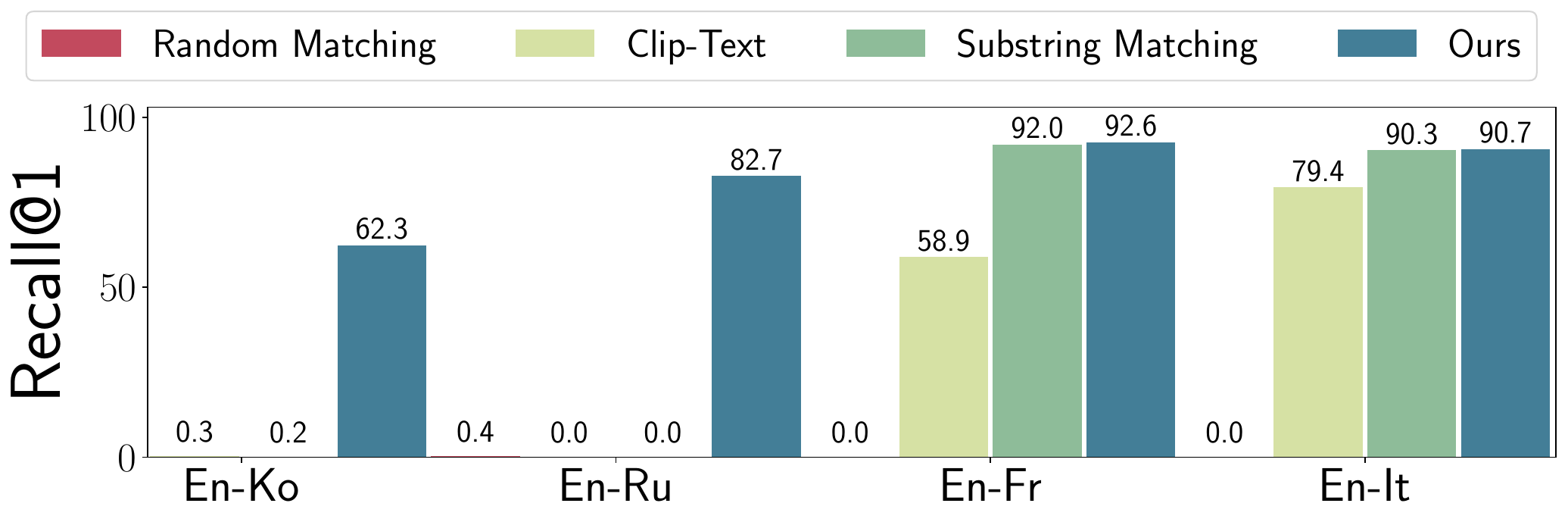}
    \vspace{-8mm}
    \caption{\ours{} with different methods of initial mapping. Compared to image-based fingerprints (dark blue), using other methods for the initial mapping result in lower recall scores, especially for dissimilar language pairs (En$\rightarrow$Ko and En$\rightarrow$Ru).
    }
    \label{fig:ablation_init}
    \vspace{-3mm}
\end{figure}

Fig.~\ref{fig:ablation_init} shows the effect of fingerprints on translation performance.
We compare variants of \ours{} using various initial mapping methods: random matching (red), clip-text embeddings (olive), substring matching (green), and image-based fingerprints (ours, dark blue).
The evaluation scores can be found in Table~\ref{tab:wiki@1}.
Fingerprint-based \ours{}s are the best among variants across all pairs.

\subsubsection{Effect of Robust \proc{}}
\begin{figure}[t]
    \centering
    \includegraphics[width=0.45\textwidth]{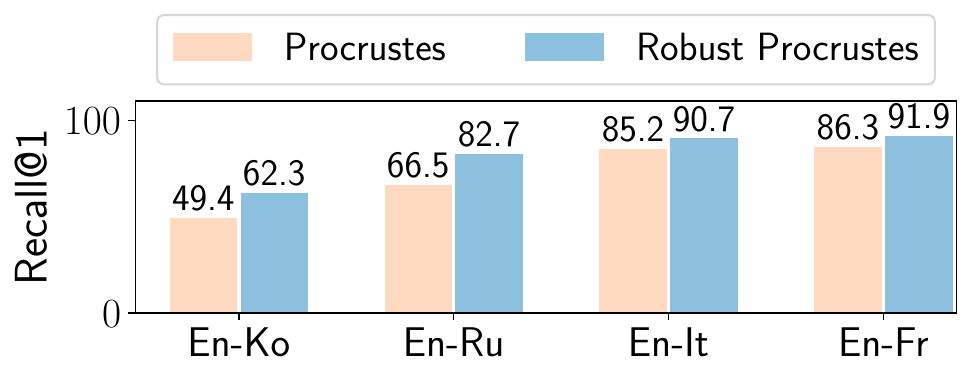}
    \vspace{-4mm}
    \caption{Investigating the effect of robust \proc{}.
    Robust \proc{} helps improve the translation across different language pairs. 
    The effect is more significant on ``difficult'' pairs, such as English-Russian.}
    \label{fig:ablation_proc}
    \vspace{-3mm}
\end{figure}

Fig.~\ref{fig:ablation_proc} shows the comparison between our robust \proc{} (in Algo.~\ref{alg:robust_procrustes}) and the standard \proc{} algorithm, given the same initial mapping.
Robust \proc{} indeed helps improve over the standard \proc{}, especially when two languages are dissimilar.
For instance, on En$\rightarrow$Ko, using robust \proc{} increases the final {recall@1} by 12.9\%.

\subsubsection{Effect of the Image Set}
\label{sec:ab_imageset}

Here we check how the images used for making fingerprints affect the performance of \ours{}.

\begin{figure}[t]
    \centering
    \includegraphics[width=0.45\textwidth]{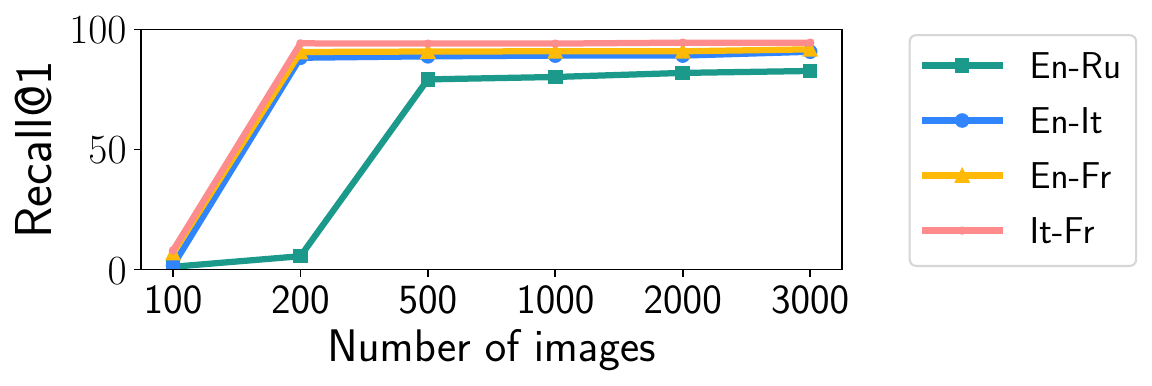}
    \vspace{-3mm}
    \caption{Recall@1 ($\uparrow$) of \ours{} varying the size of image (ImageNet) set used for fingerprints. 
    The recall improves as more images are used and remains mostly unchanged after a sufficiently large number of images (1000 to 3000 images). 
    }
    \label{fig:ablation_image}
\end{figure}

\paragraph{Size of image sets.}
Fig.~\ref{fig:ablation_image} compares {recall@1} scores of \ours{} when different number of images (from ImageNet) are used for building fingerprints.
When the number of images increases, the {recall@1} increases and converges for all pairs.
As the languages become more dissimilar, \ours{} may need more images to attain good performance.
\ours{} needs only $1000$ to $3000$ images to achieve good performance across all evaluated language pairs.

\paragraph{Diversity of images.}
\update{
To see the importance of image diversity, we fix the total number of images as $3000$ and vary the number of classes. 
Table~\ref{tab:ab_image_diversity} compares the {recall@1} of \ours{} on En$\rightarrow$Ru varying the image diversity.
Here, we use the CIFAR10 dataset for 10 or fewer classes, CIFAR100 for 20--100 classes, and ImageNet for 1000 classes.
Note that \ours{} achieve high performance only when we use a large number of classes (e.g., more than 37 classes in the Table).
This is probably because image sets with higher diversity provide more distinguishing coordinates of fingerprints to obtain more pivot pairs in the initial matching step -- the condition for robust \proc{} to learn.
Furthermore, compared to other settings, 1000-class ImageNet obtains much better initial matching in step 1.
}

\begin{table}[t]
    \footnotesize
    \sisetup{detect-weight,mode=text,group-minimum-digits =4}
    \caption{{Recall@1} ($\uparrow$) of \ours{} on  En$\rightarrow$Ru, varying the number of image classes given a fixed number of images as 3000.
    \ours{} achieves high performance (step 2) when 38 or more classes are used.
    Furthermore, using 1000-class ImageNet results in the highest initial matching score (62.2) among the settings.
    }
    \vspace{-2mm}
    \centering
    {
        \begin{tabular}{ScSS}
        \toprule[1pt]
        {No. classes} & {Dataset}  & {Step 1} & {Step 2} \\
        \hline
        1              & CIFAR10  & 0.9                  & 0.8            \\
        2              & CIFAR10  & 0.9                  & 0.6            \\
        10             & CIFAR10  & 6.2                  & 8.1            \\
        20             & CIFAR100 & 7.6                  & 5.4            \\
        37              & CIFAR100 & 9.1                & 4.4\\
        \hline
        38             & CIFAR100 & 10.3                & 82.1            \\
        50             & CIFAR100 & 11.1                 & 82.5           \\
        100            & CIFAR100 & 11.1                 & 82.3           \\
        \hline
        1000           & ImageNet & 62.2                 & 83.0          \\
        \bottomrule[1pt]
        \end{tabular}
    }
    \label{tab:ab_image_diversity}
\end{table}

\section{Conclusion}
We propose \ours{}, a novel unsupervised bilingual word alignment method using pretrained CLIP models.
\ours{} first leverages the visual similarity between words as the auxiliary for matching initial and simple word pairs via the image-based fingerprint representation computed by language-image pretraining models.
Then \ours{} uses these initial pairs as pivots to learn the linear transformation between two static word embeddings.
We introduce a robust \proc{} algorithm based on error-weighting to estimate the linear mapping.
Compared with existing baselines, \ours{} needs less computation for aligning two embeddings, %
thanks to the aid of visual information and pretrained CLIP models.
\ours{} achieves the SoTA alignment performances on several language pairs across word embedding types, especially for pairs in which two languages are highly dissimilar.
\ours{} also displays the robustness against the dissimilarity of static word embeddings' training corpora.

\newpage
\section{Limitations}
Despite achieving high translation performance on various language pairs, \ours{} has some limitations, coming from the requirements of CLIP models, the presence of visual words, and the structural similarity of static word embedding spaces.

As shown in Fig.~\ref{fig:ablation_init}, the initial mapping in Step 1 of \ours{} needs to be sufficiently good for \ours{} to achieve high translation performance.
The conditions for good initial mappings are (1) well-trained CLIP models and (2) a sufficiently large number of visual words in the two dictionaries.
\textit{First}, our setting assumes the availability of pretrained CLIP models for the two languages. 
However, this may not be the case for many languages, especially for low-resource ones having small amounts of training data publicly available.
We also observe that the CLIP models for non-English languages (either trained from scratch or fine-tuned from a model pretrained on English corpora) are not as good as the OpenAI CLIP trained on English corpora\footnote{\url{https://github.com/openai/CLIP}} in terms of image-text alignment and zero-shot image classification.
Fortunately, our results on zero-shot transfer (Fig.~\ref{fig:ablation_clip_transfer}) indicate that we may only need a few well-trained CLIP models in some major languages and further use them for their highly similar languages.
\textit{Second}, we have shown that image-based fingerprints work the best with visual words and may not show the distinguishable pattern on non-visual words (Fig.~\ref{fig:fingerprint}).
Therefore, the two dictionaries need to have a sufficient number of visual words for \ours{} to obtain initial pairs with adequate quantity and high accuracy.

\textit{Furthermore}, \ours{}, as well as most existing unsupervised word translation methods~\citep{conneau2017word,artetxe2017learning,sigurdsson2020visual} rely on the structural similarity of static word embedding spaces across languages.
However, such linear mapping between two spaces may not exist in several cases, especially when two languages are highly dissimilar.
For instance, we observed that the supervision method (with \proc{}) achieved low translation accuracy (approximately 40\%) on the English-Japanese pair evaluated on the Dictionary dataset with HTW-based embeddings, indicating that the linear transformation assumption may not be fully satisfied for these two languages' static word embedding spaces.

\section{Broader Impact and Ethical Considerations}

\ours{} provides a simple yet effective solution to word translation, contributing to the progress of machine translation, which brings more benefits to our society.
Our method is unsupervised and computationally efficient, thus significantly saving the computing and reducing the need for human labeling.
Furthermore, the robustness of \ours{} to the dissimilarity of language pairs and the dissimilarity of training corpora for static word embeddings may be beneficial to low-resource languages.

However, employing \ours{} without careful consideration and understanding may lead to undesired outcomes.
\textit{First,} the provided dictionaries may contain harmful contexts and racist or sexist content. \ours{} can be used to translate these contents to other languages, bringing unwanted adverse effects to society. 
\textit{Second}, though achieving the SOTA performances, \ours{} still has not attained sufficiently high accuracies (greater than $50\%$) on several dissimilar pairs (\textit{e.g.}, En$\rightarrow$Ja), potentially producing wrong translations for multiple words, and hence having undesired impacts to the users.
\textit{Third}, our methods may inherit biases and undesired contents from language-image (CLIP) models pretrained on large-scale datasets.
Applying efficient fine-tuning to the pretrained CLIP models with fairness consideration methods~\citep{gira-etal-2022-debiasing} may help mitigate these biases.

\newpage
\bibliography{refs}
\bibliographystyle{acl_natbib}

\appendix

\newpage

\noindent{\large \textbf{Appendix}}

\noindent Section~\ref{app:algorithms} presents the pseudocodes for algorithms discussed in the main paper.
We provide details of the experimental setting, chosen hyperparameters, computing resources, and running times in Section~\ref{sec:app_exp_setting} for reproducibility.

\section{Algorithms}
\label{app:algorithms}
We present the pseudocodes for algorithms in Section~\ref{sec:walip} of the main paper, including the \ours{} algorithm (Algo.~\ref{alg:C_UWT}), the visual-word filtering algorithm (Algo.~\ref{alg:filter-words}), and the 
word matching algorithm (Algo.~\ref{alg:csls_nn}).

\begin{algorithm}[h]
\SetNoFillComment %
\caption{\texttt{\ours{}} %
}
\label{alg:C_UWT}
\SetKwInput{Input}{Input}
\SetKwInOut{Output}{Output}
\Input{
    {Source dictionary $A_{\op{dict}} = \{a_1, \cdots, a_{n_a}\}$, 
    target dictionary $B_{\op{dict}} = \{b_1, \cdots, b_{n_b}\}$,
    CLIP models $(A^{\op{txt}}, A^{\op{img}})$, 
    $(B^{\op{txt}}, B^{\op{img}})$,
    set of images $G = \{g_1, \cdots, g_d\}$, ~~~~~~~~~~
    word vectors $T_A$ for $A_{\op{dict}}$, $T_B$ for $B_{\op{dict}}$,
    number of alignment steps $M$, threshold quantile $q$, number of candidates $k$.
    }
}
\Output{
    $\pi: [n_a] \rightarrow [n_b]$ such that $a_{\pi(i)} \equiv b_i$
}
{
\texttt{/*} \textsc{ Step 1. Pairing using fingerprints}  \texttt{*/}\\
}
{
\For{$\op{language~} l \in \{a, b\}$}
{	
$f(l_i) \leftarrow $ fingerprint in  (\ref{eqn:fingerprint}) for  $i \in [n_l]$
 }
 }
 {
$\gF \leftarrow \{f(l_i)\}_{l \in \{a, b\}, i \in [n_l]}$ \\
}
{
$\gF \leftarrow \texttt{Visual-Word-Filtering}(\gF)$ \\
}
{
$\pi_0 \leftarrow$ \texttt{Matching-Filtering}$(\gF, q)$ \\
}
{
\texttt{/*} \textsc{ Step 2. Iterative robust Procrustes}  \texttt{*/}\\
}
Set $Q_s = \{0.5, 0.5, 0.3, 0.1\}, K_s = \{10, 5, 3, 1\}$\\
Set $q = 0.5$, $k=10$, $\epsilon_0 =\infty$, $\delta = 0.5$ \\ 
\For{$m \in \{1, \cdots, M\}$}{
{
~~$s_{m-1}^A \leftarrow \{i \in [n_a]: \pi_{m-1} (a_i) \in B_{\op{dict}}  \}$ 
\\
$s_{m-1}^B   \leftarrow  \{j \in [n_b]: \hspace{-0.5mm} \exists a_i  \text{ s.t. }  \pi_{m-1} (a_i)\hspace{-0.5mm} =\hspace{-0.5mm} b_j \}$ 
\\
$T'_A \leftarrow T_A[s_{m-1}^A], \quad \quad  T'_B \leftarrow T_B[s_{m-1}^B]$\\
$W^{\star} \leftarrow$ \texttt{Robust-Procrustes}($T'_A, T'_B$)\\
$T_A \leftarrow T_AW^{\star}$ \\
$\epsilon_m = \|T_A - T_B\|_F$ \\
\If{$\epsilon_m < \epsilon_{m-1} + \delta$ }{
{
$t \leftarrow \min\{\lceil M/10 \rceil, 4\}$\\
$q \leftarrow Q_s[t], \quad k \leftarrow K_s[t]$\\
$\pi_m \leftarrow \\
~~~~~~\texttt{Matching-Filtering}(\{T_A, T_B\}, q, k)$%
}}
}
}
$\pi \leftarrow \texttt{Matching-Filtering}(\{T_A, T_B\}, 0, 1)$
\end{algorithm}

\begin{algorithm}[ht]
\SetKwInput{Input}{Input}
\SetKwInOut{Output}{Output}
\Input{Fingerprints $\gF = \{ f(l_i)\}_{l \in \{a, b\}, i\in [n_l]} $}
\Output{Updated fingerprints $\gF$ }
\For{$l \in \{a, b\}$}{
{
{
    $f_{i,j}^{(l)}\leftarrow$ $j$-th element of $f(l_i)$, for $j \in [d]$\\ 
    $f_{i, \op{max}}^{(l)} \leftarrow \max_j f_{i,j}^{(l)}$ for $i \in [n_l]$ \\
    $S_l \leftarrow \{i : f_{i, \op{max}}^{(l)} \geq \op{median}_i (f_{i, \op{max}}^{(l)}) \}$ \\
 }
\For{$i \in S_l$}{
{
$\bar{q} \leftarrow \texttt{0.9-th quantile of }\{f_{i,k}^{(l)}\}_{k=1}^d$ \\
$f_{i,j}^{(l)} \leftarrow f_{i,j}^{(l)} \cdot \mathbf{1}_{ \{ f_{i,j}^{(l)} \geq \bar{q} \} } $\\
$f_{i,j}^{(l)} \leftarrow f_{i,j}^{(l)} / \lvert f_{i,j}^{(l)} \rvert $
}
}
}
}
\caption{$\texttt{Visual-Word-Filtering}$}
\label{alg:filter-words}
\end{algorithm}

\begin{algorithm}[ht]
\SetKwInput{Input}{Input}
\SetKwInOut{Output}{Output}
\Input{{
    $\gF = \{ f(l_i)\}_{l\in \{a, b\}, i\in [n_l]}$,\\
    ~~~~~~~~~~~~~Threshold quantile $q$,\\ ~~~~~~~~~~~~~Number of candidates $k$ ($k=1$ by default).
}}
\Output{ Word index mapping $\pi: [n_a] \rightarrow [n_b]$ 
}
{
$c_{i,j} \leftarrow \op{CSLS}(f(a_i), f(b_j))$ for $i \in [n_a], j \in [n_b]$ \\
$\bar{c} \leftarrow \texttt{q-th quantile of } \{c_{i,j}\}$
\\
$\pi \leftarrow$ empty mapping from $[n_a]$ to $[n_b]$ \\
}
\For{$i \in [n_a]$}{
{
$J^{\star} \leftarrow \{j \in [n_b]: c_{i,j} \geq k\op{-th} \max_j c_{i,j} \}$\\
}
$\pi(i) \leftarrow \{j \in J^{\star} : c_{i,j} \geq \bar{c} \}$
}
\caption{$\texttt{Matching-Filtering}$}
\label{alg:csls_nn}
\end{algorithm}

\section{Experimental Setup, Implementation, and Running}
\label{sec:app_exp_setting}
We present details of the experimental setting (Sec.~\ref{sec:exp_setting} in main paper) and the chosen hyperparameters in (\ref{sec:app_exp_setup}), the computing sources, running time, and validation performance in (\ref{sec:app_exp_training}).

\subsection{Experimental Setup}
\label{sec:app_exp_setup}
\paragraph{Static word embeddings.}
We use two embeddings: \htw{} (HTW)-based Word2Vec~\citep{miech2019howto100m,sigurdsson2020visual} that trains Word2Vec~\citep{mikolov2013distributed} on HTW video datasets and Wiki-based Fasttext~\citep{bojanowski2016enriching} that trains Fasttext on the Wikipedia corpus.

\paragraph{Evaluation benchmark and datasets.} We use the \textit{Dictionary} datasets~\citep{sigurdsson2020visual} which are test sets of \muse{} bilingual dictionaries~\citep{conneau2017word}.
Each test set provides a set of matched pairs in two languages where each word in the source language can have multiple translations in the target language.
For instance, the En$\rightarrow$Fr dictionary has 1500 unique English words and 2943 corresponding French words.
All pairs used in our evaluation are En$\rightarrow$\{Fr, Ru, It, Ko, Ja\}, and It$\rightarrow$Fr.
Input evaluation dictionaries are pre-processed to ensure the delimiting character is a white-space character and that there are no duplicate synonym pairs. Words that do not appear in the word2vec files for \htw{}-based or  Wiki-based embeddings were removed.
We also provide the modifications of the original datasets that remove \textit{non-native} words (\textit{e.g.,} 'dot, gif' in the Korean dictionary).
We provide all evaluated datasets in our source codes.
The test dictionaries can also be found at \url{https://github.com/facebookresearch/MUSE} and \url{https://github.com/gsig/visual-grounding/tree/master/datasets}.

\paragraph{Evaluation metrics.} Our metric is \textit{recall@n} used in~\cite{sigurdsson2020visual} for $n=1, 10$: the retrieval for a query is correct if at least one of $n$ retrieved words is the correct translation of the query.
\textit{Recall@n} presents the fraction of source words that are correctly translated.
In our setting, the \textit{recall@1} is equivalent to \textit{precision@1}, and the matching accuracy used in~\citep{conneau2017word}.

\paragraph{Baselines.} 
We describe what baselines we have compared in this paper.
\textbf{CLIP-NN} is a simple baseline that performs double $1$-nearest neighbor ($1$-NN) on \lip{} embeddings: 
Given a source word, we perform the $1$-NN to find the nearest image (using source CLIP) and then perform the $1$-NN on the target CLIP to find the nearest target word.
For \textit{recall@n}, we perform the similar double $k$-NN where $k=\lceil\sqrt{n}\rceil$.
\textbf{MUSE}~\citep{conneau2017word} is a text-only method that learns the cross-lingual linear mapping  via adversarially aligning embeddings' distributions and iterative refinement with \proc{}.
As the adversarial training is sensitive to initialization, we follow the procedure in~\citep{sigurdsson2020visual} and report the highest observed performance across different initializations on the test set. As a result, this represents an upper bound on the true performance of the baseline.
\textbf{MUVE}~\citep{sigurdsson2020visual} replaces the linear layer learned in the first stage of \muse{} with the \textit{AdaptLayer} learned by jointly training the embeddings of videos and captions, shared across languages.
The \textit{AdaptLayer} allows monolingual embeddings to be transformed into a shared space so the rest of the network can be shared, even if the input languages are different.
Their results suggest that visually grounding translation with video allows for more robust translation.
We use their reported performances~\citep{sigurdsson2020visual} in our comparison.
\textbf{Globetrotter}~\citep{suris2020globetrotter} uses image-caption pairs to jointly align the text embeddings of multiple languages to image embeddings using a contrastive objective. 
Even though their model was trained on pairing sentences with images, they show that the text representation learned by their model can also be used for unsupervised word translation by using the Procrustes algorithm on the learned word embeddings. 
We use their word embeddings for word translation.
We also evaluate the \textbf{supervision} method using the \proc{} on different ground-truth translation pairs and use its results as an upper bound of performance.

\paragraph{Implementation details.} Here, we provide the details for implementing our algorithms. \\
\textbf{CLIP models.} We use publicly available pretrained CLIPs for English\footnote{\url{https://github.com/openai/CLIP}}, Russian\footnote{\url{https://github.com/sberbank-ai/ru-clip}}, Korean\footnote{\url{https://github.com/jaketae/koclip}}, and Japanese.\footnote{\url{https://huggingface.co/rinna/japanese-clip-vit-b-16}}
For other languages, we fine-tune English CLIP models on Multi30K~\citep{W16-3210,elliott-EtAl:2017:WMT} and MS-COCO datasets~\citep{lin2014microsoft,IJCOL:scaiella_et_al:2019,MS-COCO-ES} with translated captions for each target language.
Precisely, we fine-tune each model for $20$ epochs using the \texttt{NCEInfo} loss~\citep{oord2018representation} without changing the architectures of the original CLIP's encoders.
We use Adam optimizer~\citep{kingma2014adam} ($\beta_1,\beta_2=0.9, 0.98$) with a learning rate of \texttt{1e-7} and cosine annealing scheduler~\citep{loshchilov2016sgdr}.\\
\textbf{Image datasets.} We use 3000 images from  ImageNet~\citep{deng2009imagenet}.
We find that high-resolution images provide the best initial mappings among tested image data.\\
\textbf{Prompts for words in CLIPs.} As for the input of CLIPs, we convert every single word to a complete sentence.
We use the prompt templates suggested in~\citep{radford2021learning} and apply prompt-ensemble~\citep{radford2021learning} for the best embedding.
In particular, we use a set of (two to seven) prompts for each word and average these text embeddings as the word embedding.

\paragraph{Hyper-parameters.} 
The robust \proc{} algorithm (Algo.~\ref{alg:robust_procrustes}) uses $M=5$ iterations.
In Algo.~\ref{alg:C_UWT}, we use $M=40$ alignment iterations in Step 2 and select the best model by our evaluation loss.
We observe that the evaluation losses on pairs of similar languages (\textit{e.g.}, English-French) converge quickly after a few iterations, while the dissimilar pairs require more iterations.
For quantile threshold $q$, we use the simple scheme by gradually reducing $q$ from a set of discrete values $\{0.7, 0.5, 0.3, 0.1\}$.
The number of candidates $k$ is decayed using the following values  $\{10, 5, 3, 1\}$.

\subsection{Computation and Evaluation of \ours{}}
\label{sec:app_exp_training}

\begin{table}[t]    
    \footnotesize
    \renewrobustcmd{\bfseries}{\fontseries{b}\selectfont}
    \sisetup{detect-weight,mode=text,group-minimum-digits =4}
    \caption{Estimated \ours{} validation loss (Euclidean distance) on several language pairs performed on the HTW-based embedding and Dictionary dataset. }
    \vspace{-2mm}
    \centering
    \resizebox{0.8\linewidth}{!}
    {
        \begin{tabular}{cSSSSSS}
        \hline
        \toprule[1pt]
        \multirow{1}{*}{} & \multicolumn{1}{c}{En$\rightarrow$Fr} & \multicolumn{1}{c}{En$\rightarrow$Ko} & \multicolumn{1}{c}{En$\rightarrow$Ja}\\
        \hline
         Avg. Dist. & 8.49 & 8.58 & 8.55\\
        \toprule[1pt]
        \end{tabular}
    }
    \label{tab:val_htw}
    \vspace{-1mm}
\end{table}

\begin{table}[t]    
    \footnotesize
    \renewrobustcmd{\bfseries}{\fontseries{b}\selectfont}
    \sisetup{detect-weight,mode=text,group-minimum-digits =4}
    \caption{Estimated \ours{} validation loss (Euclidean distance) on several language pairs performed on the Wiki-based embedding and Dictionary dataset. }
    \vspace{-2mm}
    \centering
    \resizebox{\linewidth}{!}
    {
        \begin{tabular}{cSSSSSS}
        \hline
        \toprule[1pt]
        \multirow{1}{*}{} & \multicolumn{1}{c}{En$\rightarrow$Ko} & \multicolumn{1}{c}{En$\rightarrow$Ru} & \multicolumn{1}{c}{En$\rightarrow$Fr} & \multicolumn{1}{c}{En$\rightarrow$It} \\
         Avg. Dist. & 15.70 & 14.24 & 10.99 & 11.67\\
         \hline
         \multirow{1}{*}{} & \multicolumn{1}{c}{En$\rightarrow$Es} & \multicolumn{1}{c}{En$\rightarrow$De} & \multicolumn{1}{c}{Es$\rightarrow$De} & \multicolumn{1}{c}{It$\rightarrow$Fr} \\ 
          Avg. Dist. & 10.79 & 12.06 & 13.28 & 11.54 \\
         \hline
        \toprule[1pt]
        \end{tabular}
    }
    \label{tab:val_fasttext}
    \vspace{-1mm}
\end{table}

\paragraph{Validation.} As \ours{} is unsupervised, we estimated the validation error (or loss) by evaluating the average squared Euclidean distance between the mapped source embeddings and the target word embeddings. 
We use this criterion to select our best mappings.
We report the validation errors in Table~\ref{tab:val_htw} and Table~\ref{tab:val_fasttext} for evaluated language pairs on two types of static word embeddings.
We can see that the validation errors of the dissimilarity of language pairs (\textit{e.g.,} En$\rightarrow$Ko) are higher than the similar pairs (\textit{e.g.}, En$\rightarrow$Fr). 
We report the \textit{recall@1} scores corresponding to mappings with the smallest validation errors.

\paragraph{Computing resources and time.} We run our algorithms and baselines on an NVIDIA GeForce RTX 3090 GPU.
The average running time of \ours{} is about less than $2$ minutes, while MUSE models take approximately an hour for each language pair.

\paragraph{Number of parameters of \ours{} models.}  Each of our pretrained CLIP models has about 150 million trainable parameters. 
In ablation studies, we have tested \ours{} with larger versions of CLIP models, with upwards of 400 million trainable parameters.
However, we find that both \ours{} versions with smaller and large CLIP models share similar translation performances across different language pairs.

\end{document}